\documentclass{article}
\usepackage{spconf,amsmath,graphicx}
\usepackage{spconf,amsmath,graphicx}
\usepackage{arabtex}
\usepackage[ruled,vlined]{algorithm2e}
\usepackage[utf8]{inputenc}
\usepackage{utf8}
\usepackage{bbding}
\usepackage{pifont}
\usepackage{wasysym}
\usepackage{amssymb}
\usepackage{xcolor}
\usepackage{url}
\usepackage{enumitem}
\ninept

\title{Unsupervised Data Selection for TTS: \\
Using Arabic Broadcast News as a Case Study}
\name{
\begin{tabular}{c}
Massa Baali$^{1}$, Tomoki Hayashi$^{2,3}$, Hamdy Mubarak$^{3}$, Soumi Maiti$^{4}$, \\ 
Shinji Watanabe$^{4}$, Wassim El-Hajj$^{5}$, Ahmed Ali$^{3}$
\end{tabular}
}
\address{$^{1}$KANARI AI , California, USA, 
$^{4}$Carnegie Mellon University, Pittsburgh, USA \\
$^{3}$Qatar Computing Research Institute, HBKU, Doha, Qatar \\
$^{2}$Nagoya University, Japan, $^{3}$Human Dataware Lab. Co., Ltd., Japan \\
 $^{5}$American University of Beirut, Computer Science Department, Beirut, Lebanon\\
}
\begin{document}
%
\maketitle
\begin{abstract}
Several high-resource Text to Speech (TTS) systems currently produce natural, well-established human-like speech. In contrast, low-resource languages, including Arabic, have very limited TTS systems due to the lack of resources. We propose a fully unsupervised method for building TTS, including automatic data selection and pre-training/fine-tuning strategies for TTS training, using broadcast news as a case study. We show how careful selection of data, yet smaller amounts, can improve the efficiency of TTS system in generating more natural speech than a system trained on a bigger dataset. We adopt to propose different approaches for the: 1) data: we applied automatic annotations using DNSMOS, automatic vowelization, and automatic speech recognition (ASR) for fixing transcriptions' errors; 2) model: we used transfer learning from high-resource language in TTS model and fine-tuned it with one hour broadcast recording then we used this model to guide a FastSpeech2-based Conformer model for duration. Our objective evaluation shows \textbf{3.9\%} character error rate (CER), while the ground truth has \textbf{1.3\%} CER. As for the subjective evaluation, where $1$ is bad and $5$ is excellent, our FastSpeech2-based Conformer model achieved a mean opinion score (MOS) of \textbf{4.4} for intelligibility and \textbf{4.2} for naturalness, where many annotators recognized the voice of the broadcaster, which proves the effectiveness of our proposed unsupervised method.
\end{abstract}
\begin{keywords}
Text-to-Speech, TTS, low resources, speech recognition
\end{keywords}
\section{Introduction} \label{sec:intro}
Great progress in deep learning and neural end-to-end approaches have lowered the barrier to build high quality TTS systems \cite{zen2019libritts,ren2019fastspeech, tachibana2018efficiently}. However, training end-to-end TTS systems requires a large amount of high-quality paired text and audio, which are expensive and time consuming to build. 
Most of the recent TTS systems require sizable amounts of data; More than $260$ hours by professional voice actors \cite{donahue2020end}, $25$ hours \cite{wang2017tacotron}, $585$ hours constructed from $2,456$ speakers \cite{zen2019libritts} and $24$ hours \cite{ren2019fastspeech, ren2020fastspeech}. Unlike the data collection in building ASR systems, TTS needs careful recording control, since the TTS performance is tied to professional speaker and high-quality sizable recordings. 
Thus where high-resource languages such as English, Japanese, and Mandarin \cite{zen2019libritts,sonobe2017jsut,hayashi2020espnet} has reached almost human-like performance in TTS, limited attention is given to under-resource languages \cite{tu2019end} or resource efficiency of TTS \cite{chung2019semi}.

In  \cite{chung2019semi}, they proposed to use semi-supervised training framework by allowing Tacotron to utilize textual and acoustic knowledge contained in large corpora. They condition the Tacotron encoder by embedding input text into word vectors. The Tacotron decoder was then pre-trained in the acoustic domain using an unpaired speech corpus. 
In \cite{tu2019end}, authors introduce cross-lingual transfer learning for low-resource languages to build end-to-end TTS. 
To tackle input space mismatch across languages, authors proposed a phonetic transformation network model by mapping between source and target linguistic symbols according to their pronunciation. Another approach for cross-lingual transfer learning presented by \cite{xu2020lrspeech} initializes the phoneme embeddings from scratch for low-resource languages and discards the pre-trained phoneme embeddings. There has been some work on data efficiency in TTS, researchers in \cite{zhang2020unsupervised} used vector-quantization variational-Autoencoder to extract the unsupervised linguistic units from large untranscribed speech then fine-tune it on labeled speech. Others \cite{gallegos2020unsupervised} described a new unsupervised speaker selection method based on clustering per-speaker acoustic representations.


In this work, we describe a fully unsupervised framework to build a low-resource TTS system in Arabic using broadcast news recordings. This is an interesting case study for low-resource TTS as we use Arabic Aljazeera Broadcast news data from QASR corpus \cite{mubarak2021qasr} as our data source. Broadcast news recordings are widely available for various languages and possibly for large number of hours~\cite{bell2015mgb,ali2016mgb}. 
Though such broadcast news data is not very high-quality as TTS studio recordings. 


In this paper, we utilize a different data-selection strategy, text processing and speech synthesis method. First, in data-selection we show that with expert human annotations we can select 1-hour subset of high-quality data. We also show with automatic annotations using DNSMOS~\cite{reddy2021dnsmos} we can also achieve very close performance to expert labelled data, character error rate (CER) of 4.0\% compared to CER of 3.9\% with manual labels. Second, in text-processing, we show that using vowelization we can reduce CER from 32.2\% to 3.9\%. Finally, in speech synthesis we show that using transfer learning from high-resource language in TTS model and using this autoregressive (AR) TTS model to guide a non-autoregressive (non-AR) TTS model for duration, we can achieve naturalness score of 4.2 and intelligibility score of 4.4 using only 1 hour data. The reproducible recipe have been shared with the community through the ESPnet-TTS project \footnote{\tiny \url{https://github.com/espnet/espnet/tree/master/egs2/qasr_tts/tts1}}.




\section{Building TTS Corpus from news recording} 
\label{sec:data}
As per our knowledge there is no standard TTS corpus for our low-resource language Arabic.
Hence, we use the MGB-2 corpus \cite{ali2016mgb} obtained from the Aljazeera Broadcast news channel, this data was initially optimized for ASR.
Collected data spans over 11 years from 2004 until 2015. It contains more than 4,000 episodes from 19 different programs covering different domains like politics, society, economy, sports, science, etc. For each episode, Aljazeera provided the following: (\textit{i}) audio sampled at 16KHz; (\textit{ii}) manual transcription; and (\textit{iii}) metadata for most of the recordings. 
Metadata contains following information: program name, episode title and date, speaker names and topics of the episode.  Also no alignment information is provided with the textual transcriptions.
The quality of the transcription varied significantly; the most challenge is conversational programs with overlapping speech and dialectal usage; and finally metadata is not always available or standardized. 


\subsection{Speaker Identification and Speaker Linking} \label{sub:speaker_id}
Majority of metadata information appears in the beginning of the file. However, some of them are embedded inside the episode transcription. One of the main challenges is the inconsistency in speaker names, e.g. \textit{Barack Obama} appeared in $9$ different forms (\textit{Barack Obama}, \textit{Barack Obama/the US President}, \textit{Barack Obama/President of USA},etc.). 
The list of guest speakers and episode topics are not comprehensive, with many spelling mistakes in the majority of metadata field names and attributes. To overcome these challenges, we applied several iterations of automatic parsing and extraction followed by manual verification and standardization. This data is publicly available \footnote{\url{https://arabicspeech.org/qasr_tts}}.

\subsection{Data Selection for TTS Corpus}
\label{sec:dataTTS}

Our next step in building the TTS corpus is selecting anchor speakers from the MGB-2 dataset. 
We picked recordings from two anchors as our main speakers, one male and one female. Intuitively, anchor speakers will be speaking clearly, and most of the time, their recordings will be done in a high-quality studio with less noisy environment. Unlike an interviewee over the phone or an audio reporter covering events outside the studio. By investigating some recordings, we can identify the following six classes and only the last one is our desired data: 
\begin{itemize}[noitemsep,topsep=0pt,parsep=0pt,partopsep=0pt]
    \item Background music normally happens at the beginning or at the end of each episode;
    \item Wrong transcription often happens due to the fact that broadcast human transcription is not verbatim, e.g. sometimes some spoken words were dropped due to repetition, correction or rephrasing;Manual dataset classification
    \item Overlap speech occurs mainly in debates and talk-shows;
    \item Wrong speaker label when there is a problem in the meta-data (speaker linking results);
    \item Bad recording quality happens when the anchor speaker reports from a noisy environment, far-field  microphone, or calling over the phone..etc; and
    \item \textbf{Good recording}, none of the above. 
\end{itemize}
\subsubsection{Manual dataset classification}
To classify the good recordings we hired a professional linguist to listen to all segments from the selected best two anchor speakers and classify them into these aforementioned six-classes. The biggest two challenges are bad recordings (40\%) and overlapped speech (33\%). Out of this task, we were able to manually label about one hour per speaker, which we studied for building a TTS system. We ran experiments on these two different datasets mainly to assess the quality of the proposed recipe in the experimental part.

\subsubsection{Automatic dataset classification}
\label{sec:mos}
The second approach we investigated to choose the good samples is automatic data classification because manual labelling is time consuming. We use three different methods MOSNet, wv-MOS, and DNSMOS. 
\begin{itemize}[noitemsep,topsep=0pt,parsep=0pt,partopsep=0pt]

\item \textbf{MOSNet \cite{mosnet}} is a deep learning-based assessment model to predict human ratings of converted speech. They adopted the convolutional and recurrent neural network models to build a mean opinion score (MOS) predictor.  
\item \textbf{wv-MOS \cite{andreev2022hifi++}} Due to the limitations of MOSNet’s ability to detect obvious distortions they trained a modern neural network model wav2vec2.0 on the same data as MOSNet. They found that the wv-MOS was an effective predictor of subjective speech quality. 
\item \textbf{DNSMOS \cite{reddy2021dnsmos}}  is a Convolutional Neural Network
(CNN) based model that is trained using the ground truth human
ratings obtained using ITU-T P.808 \cite{naderi2020open, reddy2020interspeech}. It can be applied to stack rank various Deep Noise Suppression (DNS) methods based on MOS estimates with great accuracy. 
\end{itemize}
 Our input to the proposed methods are the wav files and the output is MOS score ranging from 1 to 5, with lowest score of 1 and highest score of 5. We considered all the MOS scores that are above 4 as excellent samples. In Table \ref{tab:recordings_dist} we shows various challenges in random 10 hours for the selected two anchor speakers with manual selection. Additionally, we present the labeled good segments from the automatic selection. We investigated the MOS score prediction for each approach by analyzing the common good samples chosen by the manual annotators. We found that DNSMOS \& wvMOS achieved high correlation with human annotators ratings for the good segments. However, such MOS rating evaluations solely took into account the quality of the speech, regardless of any transcription errors. In order to consider the good samples with wrong transcriptions and achieve our goal of building a fully unsupervised TTS, we used the Arabic ASR in \cite{hussein2021arabic} to fix the spelling mistakes.
\begin{table} [!ht]
\centering
\caption{{\footnotesize \textit{Details of the TTS. Segments(Seg.) and Duration (Dur.) in minutes for each class }}}
\scalebox{0.80}{
\begin{tabular}{lllll}
\hline 
& \multicolumn{2}{c}{\textbf{Male}}& \multicolumn{2}{c}{\textbf{Female}}\\
\textbf{Class} & \textbf{\# Seg.} & \textbf{Dur.} & \textbf{\# Seg.} & \textbf{Dur.}\\ \hline
    Background music & 631 & 70 & 1,586 & 131 \\ 
    Wrong transcription & 134 & 15 & 930 & 62  \\
    Overlapped speech  & 1,860 & 200 & 4,028 & 260  \\
    Wrong speaker & 22 & 3 & 292 & 12 \\
    Bad recordings & 2,200 & 240 & 1,312 & 105 \\
	\textbf{Good Segments} & \textbf{1,200} & \textbf{60} &  \textbf{1,094} & \textbf{70} \\ 
\textbf{DNSMOS} & \textbf{687} & \textbf{53} &  \textbf{1,241} & \textbf{97} \\ 
 MOSNet & 1209 & 70 & 1334 & 87  \\
 wv-MOS & 1315 & 63 & 933 & 44 
 \\ 
 \hline
\end{tabular}
}
\label{tab:recordings_dist}
\end{table}
\subsubsection{Vowelized Text}
\label{sec:vowelData}
Our text was without vowelization (aka diacritization) which makes the problem very challenging. In order to diacritize a text, a complex system that considers linguistic rules and statistics is needed \cite{abdelali2016farasa}. Furthermore, there are  differences between diacritizing a transcribed speech and a normal written text such as correction, hesitation, and repetition. That further needs more attention from the current text diacritizers. Additionally, the diacritized text should match the speakers' actual pronunciation of words even if they are not grammatically correct. In the experimental part, we will see the impact of the vowelization on affecting the quality of the produced results. 



\section{Architecture} \label{sec:arch}

In this section, we will introduce the architecture of text-to-mel models and vocoder models used in the evaluation.
We employed AR and non-AR text-to-mel models to compare the performance with a small amount of training data.

\subsection{Autoregressive models} \label{subsec: adapt}
We trained Tacotron2~\cite{shen2018natural} and Transformer-TTS~\cite{li2019neural} as our AR text-to-mel model. 
Tacotron2, is a recurrent neural network (RNN)-based sequence-to-sequence model.
It consists of a bi-directional Long short-term memory (LSTM)-based encoder and a unidirectional LSTM-based decoder with location sensitive attention \cite{chorowski2015attention}.
After the decoder, the convolutional PostNet refines the predicted sequence by predicting a residual component of the target sequence.
Transformer-TTS adopts a multi-headed self-attention mechanism by replacing the RNNs with the parallelizable self-attention structure. This enables faster and more efficient training while maintaining the high perceptual quality comparable to the Tacotron2 \cite{li2019neural}. 
We additionally used the guided attention loss~\cite{tachibana2018efficiently} to help the learning of diagonal attention weights for both AR models.

Since our training data was limited, it was challenging for the model to learn the alignment between the input sequence and the target sequence from one hour of speech.
To address this issue, at first, we performed pre-training on the LJSpeech dataset \cite{ito2017lj}, which consists of 24 hours of single-female English speech.
After pre-training, we fine-tuned the pre-trained model using a small amount of Arabic training data.
We initialized the network parameters except for the token embedding layer because of the mismatch in languages.


\subsection{Non-autoregressive models}
\label{subsec: dur}
We trained FastSpeech2~\cite{ren2020fastspeech} based on Conformer~\cite{guo2020recent} as our non-AR text-to-mel model, which uses the Conformer block instead of the Transformer block.
It consists of a Conformer-based encoder, a duration predictor, a pitch predictor, an energy predictor, a length regulator, a Conformer-based decoder, and PostNet.
The duration predictor is a convolutional network that predicts the duration of each input token from the hidden representation of the encoder.
The pitch and energy predictors are convolutional networks that predict pitch and energy sequences from the encoder hidden representation, respectively.
Instead of pitch and energy sequences of the target speech, token-averaged sequences are used to avoid over-fitting ~\cite{lancucki2020fastpitch}. Their embedding is added to the encoder hidden representation. The length regulator replicates each frame of the encoder hidden representation using each input token's duration to match the time-resolution with the target mel-spectrogram. 
Finally, the decoder converts the upsampled hidden representation into the target mel-spectrogram, and PostNet refines it. 

We did not perform pre-training for FastSpeech2 as we found that the non-AR model does not require a large amount of training data. As a result we train it from scratch with much smaller training data compared to the AR model.

\subsection{Synthesis} \label{subsec: synth}
We used the Griffin–Lim algorithm (GL) \cite{perraudin2013fast} and Parallel WaveGAN (PWG) \cite{yamamoto2020parallel} to generate speech from the predicted mel-spectrogram.
In the case of GL, the sequence of the predicted mel-spectrogram was converted to a linear spectrogram with the inverse mel-basis, and then GL was applied to the spectrogram.
In the case of PWG, we trained the model from scratch with the same training data as the mel-spectrogram generation networks.
We used the ground-truth of mel-spectrogram in training while using the predicted one in inference.

\section{Experiments} \label{sec:exp}
We conducted experiments to test our proposed methodologies with LJSpeech English 24-hours of speech from single speaker, then fine-tuned with one hour in Arabic. The audio quality is evaluated by human testers. One of the common problems of E2E-TTS is that the generated speech sometimes includes the deletion and/or repetition of words in the input text due to alignment errors. To address this issue, we evaluated the word error rate (WER) and CER of generated speech using pre-trained ASR models and then automatically detected the deletion and/or repetition of words. We also experimented with the automatic data selection, as discussed in Section \ref{sec:mos} to construct the TTS corpus.

\subsection{Experimental Conditions} \label{subsec:train}
Our training process requires first training the text-to-mel prediction network on its own, followed by training a PWG independently on the outputs generated by the first network.
We investigated the performance with the combinations of the following three conditions:

\textbf{Model architecture}: To check the difference among the model architectures, we compared three architectures: Tacotron2, Transformer-TTS, and Fastspeech2, as described in Section 3. 
The AR models were pre-trained \footnote{\url{https://zenodo.org/record/4925105}} as the character-based model with the LJSpeech dataset.
We used 12,600 utterances for the training and 250 utterances for the validation in pre-training.
After pre-training, we fine-tuned the model using Arabic corpus as describe in Section \ref{sec:data}.
We used 25 utterances for development and 25 for testing and the rest for training. 
For non-AR models, we extracted the ground-truth duration of each input token from the attention weights of the teacher model with teacher forced prediction~\cite{ren2019fastspeech}.
We employed both Tacotron2 and Transformer-TTS as the teacher model and compared the performance.
Then, we trained the non-AR model from scratch using only Arabic data, with the same split of data as the AR models.

\textbf{Vowelization}: Vowelization of the input transcription is important to solve the mismatch between the input text and the output pronunciation. We performed the vowelization, as described in section \ref{sec:vowelData}. We compared the case of w/ and w/o vowelization to check the effectiveness.

\textbf{Reduction factor}: The reduction factor~\cite{wang2017tacotron} is a common parameter of text-to-mel models to decide the number of output frames at each time step. This plays an important role in achieving stable training, especially when using a limited amount of training data.
We compared the reduction factor \textbf{\textit{1}} and \textbf{\textit{3}} for each condition.

\subsection{Results and Analysis} \label{subsec:eval}
\subsubsection{Objective Evaluation}\label{subsub: objEval}
For objective evaluation, we reported the Mel-Cepstral Distortion (MCD) [dB]. We also reported Word Error Rate (WER) and CER on a single speaker using the large vocabulary speech recognition Arabic end-to-end transformer system in~\cite{hussein2021arabic}.

\begin{table} [t!]
\centering
\caption{\footnotesize \textit{Objective evaluation results for the male speaker, where ``R'' represents the reduction factor, ``Vowel.'' represents whether to vowelize the input text, and the name in paraphrases $(\cdot$) represents the teacher model used for the extraction of ground-truth durations.}}
\scalebox{0.72}{
\begin{tabular}{lcccccc}
\hline
\textbf{ID}& \textbf{Model} & \textbf{R} & \textbf{Vowel.} & \textbf{WER}[\%] & \textbf{CER}[\%] & \textbf{MCD}[dB] \\ \hline
1V &    \textit{Tacotron2} & 1 & \checkmark & 34.0 & 14.0 & 10.1 ± 2.4 \\
 1 &   \textit{Tacotron2} & 1 &  & 54.4 & 29.8 & 8.5 ± 1.4 \\
 2V &   \textit{Tacotron2} & 3 & \checkmark & 35.3 & 14.1 & 10.5 ± 2.9 \\
   2 &  \textit{Tacotron2} & 3 &  & 51.1 & 27.2 & 8.9 ± 1.6 \\ 
   3V &  \textit{Transformer-TTS} & 1 & \checkmark & 20.4 & 9.7 & 8.6 ± 1.7 \\
   3 &  \textit{Transformer-TTS} & 1 &  & 45.2 & 21.6 & 9.0 ± 1.9 \\
   4V &  \textit{Transformer-TTS} & 3 & \checkmark & 15.4 & 5.5 & 8.8 ± 0.9 \\
    4 & \textit{Transformer-TTS} & 3 &  & 19.9 & 9.6  & 8.6 ± 1.7 \\
   5V &  \textit{FastSpeech2 (Taco2)} & 1 & \checkmark & 9.5 & 4.4 & 8.3 ± 1.6 \\
    5 & \textit{FastSpeech2 (Taco2)} & 1 & & 22.8 & 7.0 & 8.5 ± 1.4 \\ 
  6V &   \textit{FastSpeech2 (Taco2)} & 3 & \checkmark & 10.8 & 4.4 & 8.3 ± 1.1 \\
  6 &  \textit{FastSpeech2 (Taco2)} & 3 &  & 26.1 & 9.6 & 8.9 ± 1.6 \\ 
  7V &   \textit{FastSpeech2 (Trans.)} & 1 & \checkmark & 9.1 & 3.9 & 8.8 ± 0.9 \\
   7 &  \textit{FastSpeech2 (Trans.)} & 1 & & 53.9 & 32.3 & 9.6 ± 1.5 \\
   8V &  \textit{FastSpeech2 (Trans.)} & 3 & \checkmark & 14.1 & 5.7 & 9.0 ± 1.0 \\
   8 &  \textit{FastSpeech2 (Trans.)} & 3 &  & 34.0 & 14.7 & 9.2 ± 1.1 \\
  - &   \textit{Ground-truth} & N/A & N/A & 3.0 & 1.3 & N/A \\
    \hline
\end{tabular}
}
\label{tab:objective_result}
\vspace{-0.1cm}
\end{table}

\begin{table} [t!]
\centering
\caption{{\footnotesize \textit{The comparison of detailed CER for the male speaker between w/ or w/o vowelization. The model was FastSpeech2 with the Transformer-TTS as the teacher model and the reduction factor was set to 1.}}}
\scalebox{0.78}{
\begin{tabular}{lccccc}
\hline
\textbf{ID} & \textbf{Model} & \textbf{Sub.} & \textbf{Ins.} & \textbf{Del.} &  \textbf{CER} [\%] \\ \hline
  7V &  \textit{w/ vowelization} & 11 & 2 & 32 & \textbf{3.9} \\
   7 & \textit{w/o vowelization} & 160 & 45 & 170 & \textbf{32.3} \\ 
 \hline
\end{tabular}
}
\label{tab:cer}
\vspace{-0.1cm}
\end{table}
The objective evaluation in Table~\ref{tab:objective_result} shows that vowelization improved results considerably. Table \ref{tab:cer} shows the impact of vowelization for the FastSpeech2 models. The generated speech in the unvowelized text compared to the vowelized text includes a lot of deletion of characters in the input text.
Usually, the model with a larger reduction factor achieves faster training convergence and easier attention alignment. However, as we can see from the results produced that it reduces the quality of the predicted frames.
1V and 3V showed reasonable performance in this particular task, but the 5V and 7V outperformed the two models even without the pre-training. 
This result indicated that the amount of training data required for non-AR models is much smaller than AR models. 

We notice that WER for the ground-truth is very low compared to any other representative test set. This can be owed to the fact that anchor speaker speaks clearly and recording setup is very good.

To check the quality of the automatic labeling done by MOSNet, wv-MOS, and DNSMOS (Section \ref{sec:mos}) vs. manual labeling provided by the annotators, we trained a FastSpeech2 model with the fine-tuned Transformers as a teacher model.
Table \ref{tab:werMoS} shows a comparison between the efficiency of the four  models trained on different samples, 
 which shows that the manual labeling slightly outperformed the automatic labeling. However the DNSMOS shows a very promising results as its slightly worse than the manual selection. 

For checking the efficiency of the data selection and the performance of the mechanism presented, we trained a FastSpeech2 model with the fine-tuned Tacotron2 as a teacher model for both the male and the female speaker dataset.
Table \ref{tab:werFemale} shows a comparison between the results of each dataset using the same model. This shows that the male speaker dataset has higher quality.

\subsubsection{Subjective Evaluation}\label{subsub: subEval}  
Finally, we conducted the subjective evaluation using MOS on naturalness and intelligibility.
We compared the following four models trained on the male dataset: 7V, 7, 8V, and 7V with PWG.
For a reference, we also used ground-truth samples in the evaluation.
We used 100 sentences of evaluation data for the subjective evaluation.
Each subject evaluated 100 samples of each model (in total 400 samples) and rated the intelligibility and naturalness of each sample on a 5-point scale: 5 for excellent, 4 for good, 3 for fair, 2 for poor, and 1 for bad.
The number of subjects was ten.

Table \ref{tab:mos} shows that with reduction factor 1,  vowelization significantly improved both intelligibility and naturalness, revealing the effectiveness of vowelization. We can confirm that a larger reduction factor did not improve the performance, although it accelerated the training speed. Finally, the result with PWG showed that applying the neural vocoder brought a significant improvement of the naturalness and the intelligibility. We can also conclude from the high results presented in the ground-truth that people are very familiar with the tone of the speakers in the broadcast news. 

\begin{table} [t!]
\centering
\caption{{\footnotesize \textit{Comparison between the manual and the automatic labeling results for the male speaker. The model was FastSpeech2 with the Transformers as the teacher model. The reduction factor was set to 1 and vowelization was performed.}}}
\scalebox{0.76}{
\begin{tabular}{lcccccc}
\hline
\textbf{ID} & \textbf{Model} & \textbf{WER} [\%]& \textbf{CER} [\%]& \textbf{MCD} [dB]\\ \hline
  10 &  \textit{MOSNet} & 22.0 & 10.0 & 12.6 ± 1.1 \\
  10V &  \textit{wv-MOS} & 18.0 & 7.0 & 10.9 ± 1.0 \\
  11 &  \textit{DNSMOS} & 11.0 & 4.0 & 11.2 ± 0.65 \\
  7V &  \textit{w/ manual labels} & 9.13 & 3.9 & 8.8 ± 0.9 \\ 
     \hline
    
\end{tabular}
}
\label{tab:werMoS}
\end{table}
\begin{table} [t!]
\centering
\caption{{\footnotesize \textit{Objective evlaution using FastSpeech2 with the Tacotron2 as the teacher model with vowelization.}}}
\scalebox{0.74}{
\begin{tabular}{lcccc}
\hline
\textbf{ID} & \textbf{Dataset} & \textbf{WER} [\%] & \textbf{CER} [\%] & \textbf{MCD} [dB]\\ \hline
   5V & \textit{Male} & 9.5 & 4.4 & 8.3 ± 1.6 \\
   11V & \textit{Female} & 13.0 & 7.0 & 7.9 ± 1.3\\ 
     \hline
    
\end{tabular}
}
\label{tab:werFemale}
\end{table}
\begin{table} [t!]
\centering
\caption{\textit{Subjective evaluation results with 95\% confidence interval, where reduction factor ``R'', intelligibility ``Int.'' and naturalness ``Nat.''.}}
\scalebox{0.70}{
\begin{tabular}{lccccc}
\hline
\textbf{ID} & \textbf{Model} & \textbf{R} & \textbf{Vowel.} & \textbf{Int.} & \textbf{Nat.} \\ \hline
  7V &  \textit{FastSpeech2 (Trans.)} & 1 & \checkmark & 4.1 ±  0.06 & 4.0 ± 0.06 \\ 
  7 &  \textit{FastSpeech2 (Trans.)} & 1  & & 3.5 ± 0.08 & 3.2 ± 0.08 \\ 
  8V &  \textit{FastSpeech2 (Trans.)} & 3 & \checkmark & 3.9 ± 0.07 & 3.8 ± 0.07 \\
  12 &  \textit{FastSpeech2 (Trans.) w/ PWG} & 1 & \checkmark  & 4.4 ± 0.06  & 4.2  ±  0.06 \\
  - &  \textit{Ground-truth} & N/A & N/A & 4.9 ± 0.05 & 4.9 ± 0.05\\ \hline
\end{tabular}
}
\label{tab:mos}
\end{table}
\section{Conclusion} \label{sec:conclusion}
This paper introduced a method for building TTS in an unsupervised way, including data selection and pre-training/fine-tuning strategies for TTS training, using Arabic broadcast news as a case study. We explained the pipeline that involves data collection, adaptation of AR models, training of non-AR models, and training PWG. Our approach resulted in MOS of 4.4/5 for intelligibility (almost excellent) and 4.2/5 for naturalness (almost very good). For future work, we plan to automate the process for all the speakers in the QASR \cite{mubarak2021qasr} corpus. Finally, we also consider applying our pipeline on another language, such as the BBC in the MGB-1.


\bibliographystyle{IEEEbib}
\bibliography{strings,refs}

\end{document}